\documentclass{article}
\usepackage{spconf,amsmath,graphicx}
\usepackage{multirow}              
\usepackage{microtype}             
\usepackage{csquotes}
\usepackage{comment}
\usepackage{xcolor}

\title{Referring Multi-Object Tracking in Moving-Camera Videos via Global Motion Compensation}
\name{Hsin-Chen Pai, Jyun-Kai Wang$^\dagger$, Yi-Cheng Peng$^\dagger$, Wei-Ta Chu}
\address{National Cheng Kung University, Tainan, Taiwan\\
  {\tt \{f74126335, f74122250, f74122137, wtchu\}@gs.ncku.edu.tw}\\
  {\small $^\dagger$ These authors contributed equally as co-second authors.}}
\begin{document}
\ninept
\maketitle
\begin{abstract}
  Referring multi-object tracking (RMOT) takes a video and a language expression as input and tracks all referred objects. Many tracking requirements involve how an object moves rather than how it appears. However, in a video captured by a moving camera, a parked vehicle may appear to move, while a moving vehicle may show little displacement. Existing RMOT methods relate motion with text but do not explicitly remove camera-induced motion. In this paper, we propose extracting residual motion across frames by estimating camera motion in driver-view videos and compare motion characteristics with the query expression. We consider the motion-matching extent and integrate it with the RMOT method's prediction result through late fusion. In the evaluation, we verify the performance gain of taking the motion compensation module as a plug-in across different RMOT hosts.
\end{abstract}


\section{Introduction}
\label{sec:intro}
Referring multi-object tracking (RMOT) outputs tracking results of the objects described by a free-form language expression. For example, a \enquote{left-vehicles-in-red} description targets the red vehicles on the left. However, many descriptions focus on moving objects, such as \enquote{moving cars} or \enquote{a vehicle turning left}. In videos captured by a moving camera, the observed motion is a mixture of object motion and camera motion; thus, a static car may appear to move, while a moving car may appear static across frames.

Previous RMOT methods~\cite{transrmot,ikun,flexhook} match language expressions to objects. VMRMOT~\cite{vmrmot} considers explicit motion cues, such as changes in direction and speed, in the matching process. These methods compute motion from the object's bounding-box displacement across frames, which mixes object motion with camera motion. On the other hand, conventional multi-object tracking (MOT) methods propose to remove camera motion. BoT-SORT~\cite{botsort} estimates camera-induced image motion, while UCMCTrack~\cite{ucmctrack} stabilizes object trajectories. Overall, how motion compensation can be extended to RMOT is still underexplored.

In this paper, we propose separating object motion from camera motion and providing motion cues for RMOT methods. It consists of four stages. First, the geometric transformation across frames is estimated to measure camera motion. The estimated transformations are accumulated across frames to reliably measure object motion. Second, residual velocity, which represents object motion after removing camera motion, is computed across frames and used as a temporal motion feature. Third, a Contrastive Motion Aligner is developed to compare motion features with language expressions and produce a motion-matching score. Finally, the motion-matching score is scaled by a host-specific weight and added to the host's matching score. Here, the host model refers to an existing RMOT method used as a baseline, such as iKUN~\cite{ikun}, FlexHook~\cite{flexhook}, and TransRMOT~\cite{transrmot}. The proposed process is implemented as a plug-in module, thereby improving existing RMOT baselines without retraining them.

This work connects motion compensation with motion-language modeling for RMOT. Our main contributions are as follows.
\begin{itemize}
  \item We propose a plug-in module that aligns the object's residual velocity with the referring expression, thereby improving the host RMOT method without modifying or retraining it.
  \item Across three host RMOT methods, i.e., iKUN, FlexHook, and TransRMOT, the proposed module improves pooled HOTA in all settings. On iKUN, it raises moving-class HOTA from $27.697$ to $37.139\pm0.923$ (approximately $+9.44$ points).
\end{itemize}


\begin{figure*}
  \centering
  \includegraphics[width=13cm]{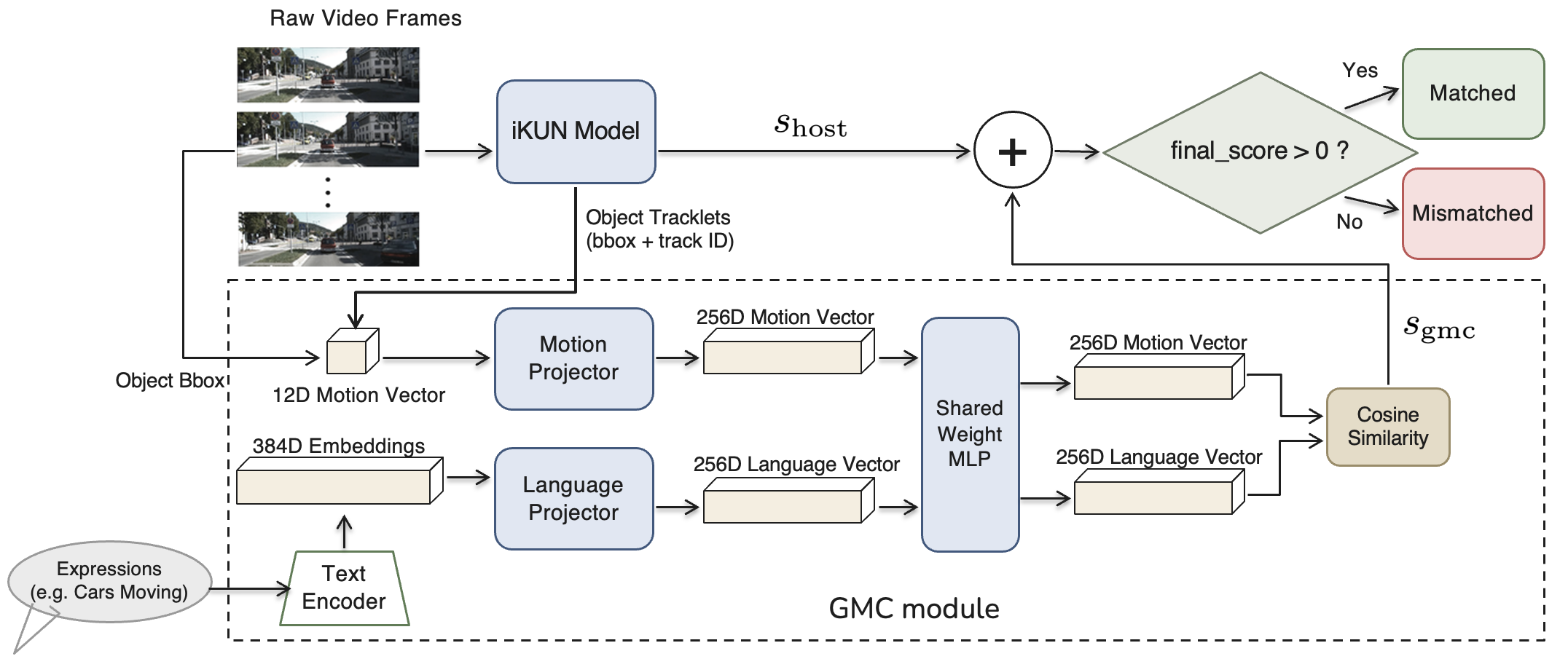}
  \caption{The proposed RMOT framework with the global motion compensation module. The host model (iKUN in this case) predicts an object-language matching score. The GMC module provides an extra matching score to adjust matching results.}
  \label{fig:architecture}
\end{figure*}

\section{Related Works}
\label{sec:related}
\subsection{Referring Multi-Object Tracking}
TransRMOT~\cite{transrmot} introduced the RMOT task on the Refer-KITTI dataset. Later methods improved different parts of this task. iKUN~\cite{ikun} separated language-based object matching from object tracking, while FlexHook~\cite{flexhook} improved how visual and language information interact. ReferGPT~\cite{refergpt} used a multimodal large language model to describe object tracks, and MEX~\cite{mex} provided a memory-efficient module for matching tracks with language.

Recent RMOT methods~\cite{deeprmot} further improve models' understanding of object descriptions. STORM~\cite{storm} unifies grounding and tracking in an end-to-end multimodal large language model. VMRMOT~\cite{vmrmot} explicitly represents motion using information such as position, direction, changes in distance, and changes in speed. These methods estimate object motion from the object's bounding-box movements across frames.

\subsection{Motion Compensation in Tracking}
Camera motion compensation is widely used in traditional multi-object tracking. BoT-SORT~\cite{botsort} estimates camera-induced motion across frames. UCMCTrack~\cite{ucmctrack} estimates how the camera moves relative to the road. EMAP~\cite{emap} further separates the camera's turning and movement from object motion within the Kalman prediction process.

These methods demonstrate that compensating for camera motion can improve tracking. However, how to integrate compensated motion cues with language expressions to improve RMOT remains underexplored.


\section{Method}
\label{sec:method}
Figure~\ref{fig:architecture} shows the proposed RMOT framework with the global motion compensation (GMC) module. The GMC Module is attached to a host RMOT method, iKUN~\cite{ikun} in this case. iKUN performs object detection and tracking and then outputs object-language matching scores. The inputs to the GMC module are the sequence of object bounding boxes across frames, the video frames, and the language expression. Based on these inputs, it estimates camera motion, extracts residual object motion, aligns it with the language expression, and produces another matching score. The score is then combined with the host model's predicted score to determine the final tracking result. The GMC module consists of four stages: motion compensation, multi-scale residual velocity extraction, motion-language alignment, and late fusion.

\subsection{Motion Compensation}
\label{sec:ego}
As in previous RMOT works, we take Refer-KITTI~\cite{transrmot} as the evaluation benchmark. Refer-KITTI is recorded from a driver-view camera, so the road surface is visible in most frames and stays close to a single plane. For a forward-facing camera the horizon lies near the vertical image center, so the lower half of the frame contains the road surface. We extract Shi-Tomasi keypoints~\cite{shitomasi} in this lower half, excluding the interiors of the tracked object boxes, and track them into the next frame with pyramidal Lucas-Kanade optical flow~\cite{lucaskanade}. The RANSAC~\cite{ransac} algorithm is then employed to estimate the frame-to-frame homography $H_{t-1 \to t}$ to describe the correspondences between keypoints in two frames.

To describe camera motion across more frames, the estimated homographies are accumulated as
\begin{equation}
  H_{t-k \to t} = H_{t-1 \to t} \times H_{t-2 \to t-1} \times \cdots \times H_{t-k \to t-k+1},
  \label{eq:cumulative}
\end{equation}
where $H_{t-k \to t}$ represents the accumulated camera transformation from frame $I_{t-k}$ to frame $I_t$. Instead of repeatedly transforming already-warped coordinates, the accumulated homography is applied only once to the detected objects' centroids. This reduces numerical drift and provides a stable reference for estimating the residual velocity in the next stage.

\subsection{Multi-Scale Residual Velocity}
\label{sec:residual}
Before going into more details, we define the following three velocities:
\begin{itemize}
  \item Raw velocity: Raw velocity is the observed image-plane velocity of an object, computed from the displacement of its centroid between two frames separated by $g$ frames and normalized by the corresponding time interval.
  \item Ego-motion-induced velocity: Ego-motion-induced velocity is the image-plane velocity caused by camera motion, computed from the displacement of the object's location predicted by the homography between two frames separated by $g$ frames and normalized by the corresponding time interval.
  \item Residual velocity: Residual velocity is the camera-motion-compensated object velocity obtained by subtracting the ego-motion-induced velocity from the raw velocity.
\end{itemize}

Based on the accumulated homography matrix and the objects' bounding boxes, we aim to estimate the \emph{residual velocity}. For a temporal gap $g$, the raw velocity $v^{\mathrm{raw}}_{g}$ is calculated by dividing the displacement of the object centroid between two frames ($I_{t-g}$ and $I_{t}$) by the time gap $g$. The raw velocity is mixed with the object's motion and the camera's motion. To remove the camera motion, the accumulated homography is applied to an object's centroid $\boldsymbol{o}_{t-g}$ at $I_{t-g}$, yielding $\hat{\boldsymbol{o}}_{t}$ that is the predicted centroid at $I_{t}$ under the assumption that the object is static. Then the ego-motion-induced displacement $\hat{\boldsymbol{o}}_{t}-\boldsymbol{o}_{t-g}$ is normalized by the time interval $g$ to estimate the ego-motion-induced velocity $v^{\mathrm{ego}}_{g}$. The residual velocity $v^{\mathrm{res}}_{g}$ is then computed as
\begin{equation}
  v^{\mathrm{res}}_{g} = v^{\mathrm{raw}}_{g} - v^{\mathrm{ego}}_{g}.
  \label{eq:residual}
\end{equation}
This value approximates the motion that a static camera would observe. For example, a parked vehicle has near-zero residual velocity even when camera motion causes a large image displacement. All displacements are normalized by the image dimensions to make them independent of the image resolution. In the implementation the factor $1/g$ is omitted; this is equivalent up to a per-gap rescaling of the first projection layer of Sec.~\ref{sec:align}.

To capture motion over different time spans, residual velocities are computed at temporal gaps of $\{2,5,10\}$ frames. That means the accumulated homography matrices mentioned in Eqn.~(\ref{eq:cumulative}) have three versions, i.e., $H_{t-2 \to t}$, $H_{t-5 \to t}$, and $H_{t-10 \to t}$. The residual velocities are concatenated with the object's box information to form a 12-dimensional motion feature,
\begin{equation}
  \bigl((dx_s,dy_s),(dx_m,dy_m),(dx_l,dy_l),\Delta w,\Delta h,c_x,c_y,w,h\bigr),
  \label{eq:descriptor}
\end{equation}
where the first six values are the residual velocities at the three temporal gaps, and $\Delta w, \Delta h$ capture the change in box scale. The terms $c_x$ and $c_y$ describe the coordinates of the box center, and $w$ and $h$ describe the width and height of the box, respectively.

The largest gap requires $11$ contiguous frames, so a track contributes a motion feature only after its history reaches that length. Shorter histories are left to the host: the module abstains and the host score passes through unchanged. The threshold is fixed by the largest gap and adds no free parameter.

\subsection{Motion-Language Matching}
\label{sec:align}
Given the motion features mentioned above and the language expression, this stage measures how much the observed object's (relative) motion matches the expression. The 12-dimensional motion feature is encoded by a motion encoder, while the language expression is encoded by the model called all-MiniLM-L6-v2 (MiniLM)~\cite{sbert} into the text embeddings. Both embeddings are projected into the same feature space using linear projection layers and a shared Multi-Layer Perceptron (MLP).

The model is trained using the Information Noise Contrastive Estimation (InfoNCE)~\cite{infonce} loss to maximize the similarity between matched motion-expression pairs while separating unmatched pairs. A positive pair is formed by the embeddings of an object and the corresponding expression. Other object-expression pairs in the same mini-batch are treated as negatives. The MiniLM text encoder is kept fixed, and only the projection layers and motion encoder are trained. During inference, the cosine similarity $s_{\mathrm{gmc}}$ between an object's motion embeddings and the target expression's embeddings is used as the matching score and passed to the late fusion stage described below.

\subsection{Fusion}
\label{sec:fusion}
We aim to integrate the matching score described above with the prediction score from the host RMOT method to improve tracking performance. The integration is defined as
\begin{equation}
  s_{\mathrm{final}} = s_{\mathrm{host}} + \alpha_{c}\,s_{\mathrm{gmc}},
  \label{eq:fusion}
\end{equation}
where $s_{\mathrm{host}}$ is the prediction score from the host model. The weighting $\alpha_c$ is designed to vary for different types of expressions. For example, \enquote{moving car} is a \emph{moving} expression, \enquote{left vehicles which are parking}\footnote{in Refer-KITTI, this expression refers to a vehicle that is already parked, rather than a vehicle that is in the process of parking.} is a \emph{static} expression, and \enquote{right cars in light color} is an \emph{appearance} expression. Based on the keywords in the expression, we classify it as moving, static, or appearance. Because different host models produce scores over different ranges, the weighting $\alpha_c$ is selected for each host via leave-one-sequence-out (LOSO) cross-validation. For each fold, the weighting that yields the highest pooled HOTA (on the validation set) is chosen for all but one sequence. The final weighting is determined as the median over folds. For both FlexHook settings, the procedure selects the same weighting for all expression classes, so the routing degenerates to a single weighting.

Since the GMC module only adds extra influence on the final prediction score, it can be integrated into existing RMOT frameworks, either two-stage or end-to-end query-based, without changing or retraining their detector, tracker, visual encoder, or language encoder.

\section{Experiments}
\label{sec:exp}

\subsection{Setup}
\label{sec:setup}
We evaluate the proposed module on the Refer-KITTI V1~\cite{transrmot} and the Refer-KITTI V2~\cite{temprmot} benchmarks using the HOTA metric from TrackEval~\cite{hota}, which jointly measures detection and association performance. We report both moving-class HOTA, which evaluates only the moving-class expressions (those assigned to the moving class by the keyword classifier of Sec.~\ref{sec:fusion}), and pooled HOTA, which evaluates all expressions in the benchmark. The V1 test split contains three sequences, and the V2 test split contains four.

The proposed GMC module is integrated with three host models, i.e., iKUN~\cite{ikun}, FlexHook~\cite{flexhook}, and TransRMOT~\cite{transrmot}. For comparison, we also evaluate a raw-velocity variant that uses image-plane motion directly, without ego-motion compensation. The GMC module is trained using the AdamW~\cite{adamw} optimizer with a learning rate of $10^{-3}$, batch size 256, and 100 epochs. The linear projectors and MLP in the module are trained based on the Refer-KITTI V1 training split (15 sequences, disjoint from the test sequences). Each positive pair consists of an expression and the motion features derived from the object's ground-truth boxes. At test time, raw object movement is derived from the host model's initial tracking results, the residual velocity is calculated, and the motion features are finally computed. The road-plane homography tracks up to 600 corners with a 3-pixel RANSAC threshold and succeeds on all 7,690 adjacent frame pairs of the 19 training and evaluation sequences. Excluding host inference, the module runs at $149$ FPS on CPU ($6.7$\,ms per frame).

Results are averaged over three random seeds for the main configurations. Ablation analyses are done using five seeds together with Welch's $t$-test. All host models use their original detection results, and every Refer-KITTI V1 setting is evaluated on the benchmark's official 150-expression test list. Our reproduced FlexHook baselines match their published scores exactly at $\alpha_c{=}0$. TransRMOT's published scores ($46.56$ before and $38.06$ after the community frame-numbering correction) use different ground-truth conventions and are not directly comparable; Table~\ref{tab:main} therefore lists our reproduction under the host's own frame convention, which the fused model reproduces exactly at $\alpha_c{=}0$.

Refer-KITTI does not provide a dedicated validation split. All fusion weights are therefore selected by the leave-one-sequence-out procedure mentioned in Sec.~\ref{sec:fusion}. No weight is chosen on the sequence it is evaluated on. The selected weightings for the four host settings are listed in Table~\ref{tab:alphas}.

\begin{table}[b]
  \centering
  \caption{Fusion weightings $\alpha_c$ selected by leave-one-sequence-out cross-validation.}
  \label{tab:alphas}
  \footnotesize
  \begin{tabular*}{\linewidth}{@{\extracolsep{\fill}}lcc}
    \hline
    Host        & $\alpha_{\mathrm{mot}}$ (moving class) & $\alpha_{\mathrm{oth}}$ (other class) \\
    \hline
    iKUN        & 1.0 & 0.1 \\
    TransRMOT   & 0.7 & 0.5 \\
    FlexHook V1 & 7   & 7   \\
    FlexHook V2 & 5   & 5   \\
    \hline
  \end{tabular*}
\end{table}

Refer-KITTI is highly imbalanced: only $21$ of the $150$ V1 and $111$ of the $862$ V2 test expressions are moving-class expressions. Consequently, improvements in motion understanding may have only a limited impact on pooled HOTA. Therefore, moving-class HOTA is used as the primary evaluation metric, while pooled HOTA is reported to reflect overall benchmark performance.

\subsection{Main results}
\label{sec:main}
Table~\ref{tab:main} summarizes the pooled HOTA results. We evaluate the performance variations when iKUN, TransRMOT, or FlexHook is used as the host model, with or without the GMC module. Table~\ref{tab:main} shows that the proposed GMC module provides performance gains across all settings. For iKUN and both FlexHook settings, the fused model also exceeds the host's published score; for TransRMOT, the comparison is against our reproduction (Sec.~\ref{sec:setup}).

\begin{table}[t]
  \centering
  \caption{Pooled HOTA, DetA, and AssA on the Refer-KITTI test sets. Each host row gives the baseline (HOTA as published; DetA and AssA from our reproduction); the +\,GMC row gives the fused model (mean over three seeds). The better value of each pair is in bold. $^*$ denotes our reproduction using the official implementation, under the host's own frame convention (Sec.~\ref{sec:setup}).}
  \label{tab:main}
  \footnotesize
  \begin{tabular*}{\linewidth}{@{\extracolsep{\fill}}lccc}
    \hline
    Method                      & HOTA               & DetA  & AssA  \\
    \hline
    \multicolumn{4}{c}{\textit{Refer-KITTI}} \\
    \hline
    iKUN~\cite{ikun}            & 44.56              & 32.04 & 62.48 \\
    \quad +\,GMC                & \textbf{45.304} & \textbf{32.90} & \textbf{62.94} \\
    \hline
    TransRMOT~\cite{transrmot}  & 45.757$^*$         & 36.64 & 57.40 \\
    \quad +\,GMC                & \textbf{46.149} & \textbf{36.95} & \textbf{57.89} \\
    \hline
    FlexHook V1~\cite{flexhook} & 53.824             & 43.35 & 66.92 \\
    \quad +\,GMC                & \textbf{53.980} & \textbf{43.51} & \textbf{67.07} \\
    \hline
    \multicolumn{4}{c}{\textit{Refer-KITTI-V2}} \\
    \hline
    FlexHook V2~\cite{flexhook} & 42.526             & 30.63 & 59.19 \\
    \quad +\,GMC                & \textbf{42.625} & \textbf{30.75} & \textbf{59.23} \\
    \hline
  \end{tabular*}
\end{table}

Table~\ref{tab:deficit} reports HOTA variations on the moving-class expressions only. As can be seen, the GMC module provides performance improvement across all four settings. The performance gains differ by an order of magnitude across the four settings. The gain decreases monotonically as the host's baseline moving-class HOTA increases, across three different architectures: the weaker the host's own motion understanding, the more the compensated motion cues contribute.

\begin{table}[t]
  \centering
  \caption{Moving-class HOTA (moving-class expressions: $21/150$ for iKUN, TransRMOT, and FlexHook V1, $111/862$ for FlexHook V2), $n{=}3$ seeds. Rows are ordered by baseline moving-class HOTA.}
  \label{tab:deficit}
  \small
  \setlength{\tabcolsep}{2pt}
  \begin{tabular*}{\linewidth}{@{\extracolsep{\fill}}lcc}
    \hline
    Host        & Baseline & Performance gain \\
    \hline
    iKUN        & 27.70    & $+9.44\pm0.92$   \\
    TransRMOT   & 33.88    & $+1.43\pm0.13$   \\
    FlexHook V2 & 38.56    & $+0.69\pm0.04$   \\
    FlexHook V1 & 47.90    & $+0.43\pm0.19$   \\
    \hline
  \end{tabular*}
\end{table}

\subsection{Ablation Studies}
\label{sec:ablation}
Table~\ref{tab:ablation} reports the five-seed ablation measurements when iKUN is used as the host model. The full module improves moving-class HOTA from $27.70$ to $36.99$ and pooled HOTA from $44.54$ to $45.28$. If the ego-motion compensation is removed, the moving-class HOTA drops from $36.99$ to $32.37$, while removing the multi-scale gaps when calculating residual velocity reduces it from $36.99$ to $35.14$. Both drops are significant: Welch's $t$-test between the five seeds of the full module and the five seeds of each variant gives $t{=}13.1$ on moving-class and $t{=}13.2$ on pooled HOTA when the ego-motion compensation is removed, and $t{=}5.3$ and $t{=}6.8$ when the multi-scale gaps are removed (all $p{<}0.01$). The expressions outside the moving class change little across the variants (Others in Table~\ref{tab:ablation}). These show the influence of both components.

\begin{table}[t]
  \centering
  \caption{Per-class HOTA ablation on iKUN ($n{=}5$ seeds); single $\alpha$: $\alpha_{\mathrm{mot}}{=}\alpha_{\mathrm{oth}}{=}0.35$ (LOSO). Others $=$ all expressions outside the moving class.}
  \label{tab:ablation}
  \footnotesize
  \setlength{\tabcolsep}{2pt}
  \begin{tabular*}{\linewidth}{@{\extracolsep{\fill}}lccc}
    \hline
    Variant                    & Moving HOTA               & Others HOTA               & Pooled HOTA               \\
    \hline
    Native host (w/o GMC)      & 27.70          & 45.87          & 44.54          \\
    Full module (w. GMC)      & \textbf{36.99} & \textbf{46.05} & \textbf{45.28} \\
    single $\alpha$ (no routing) & 32.42        & 45.99          & 44.95          \\
    $-$ ego (raw velocity only) & 32.37         & 45.77          & 44.61          \\
    $-$ multiscale (single gap) & 35.14         & 45.96          & 45.00          \\
    \hline
  \end{tabular*}
\end{table}

\begin{figure}[t]
  \centering
  \includegraphics[width=\linewidth]{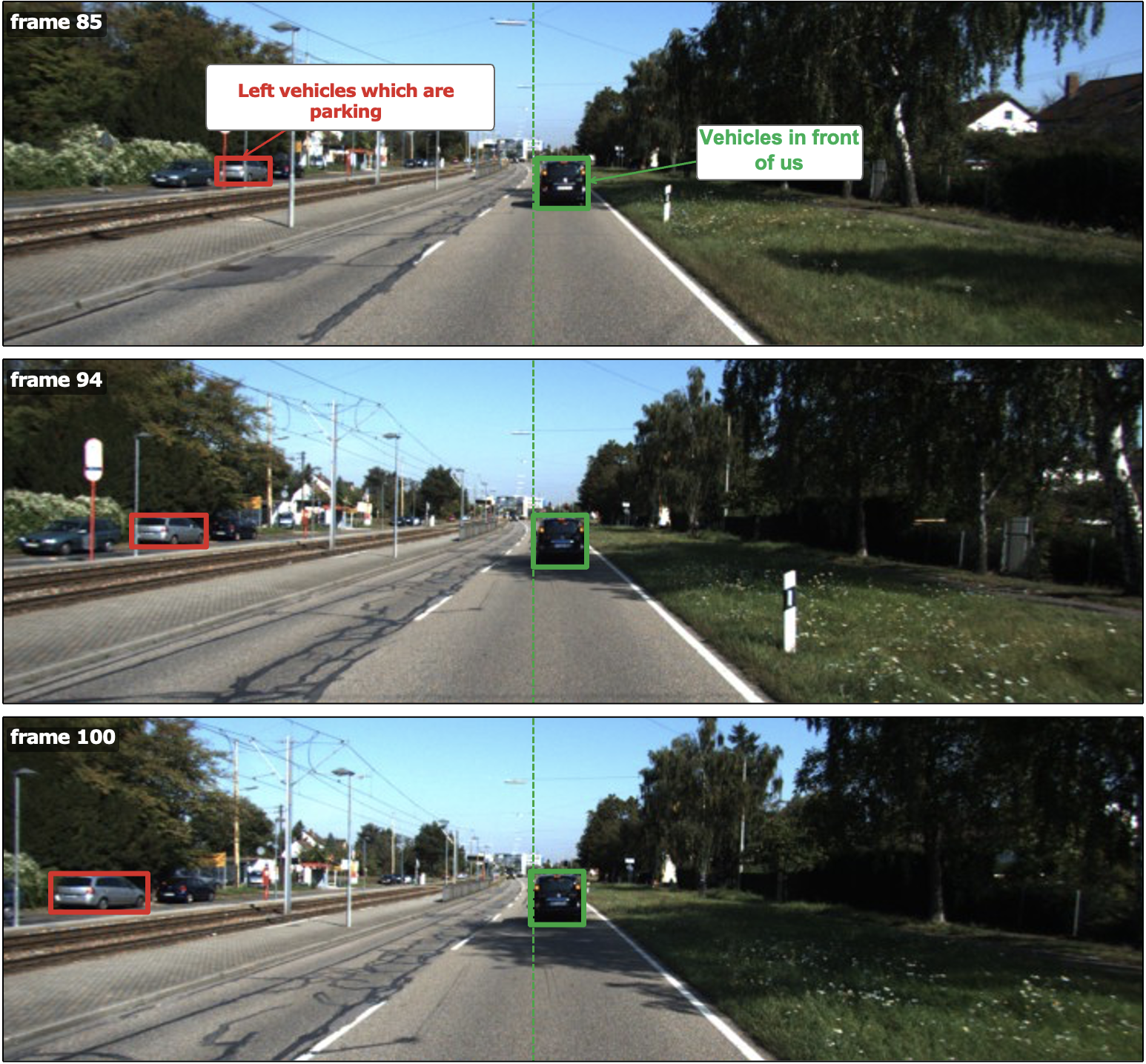}
  \caption{``Moving cars'' (Refer-KITTI seq 0005). The parked car (red) sweeps across the image as the camera passes, while the moving car (green) barely shifts. Raw image motion misclassifies both; the ego-compensated residual recovers both from the same host score.}
  \label{fig:qual}
\end{figure}

Figure~\ref{fig:qual} shows a sample tracking result. The parked car appears to move across frames due to camera motion, while the moving car remains at nearly the same distance from the camera. Without ego-motion compensation, these two cases are easily confused and mistracked. After motion compensation, the parked car has nearly zero residual velocity, while the moving car presents nonzero residual velocity. This allows the GMC module to more correctly predict the matching score.

\section{Conclusion and Limitations}
\label{sec:conclusion}
The GMC module is a plug-in that improves motion grounding in referring multi-object tracking without modifying the detector, tracker, or appearance-language matching model. By compensating for camera motion and matching residual motion features with language expressions, the proposed module helps distinguish object motion from camera-induced motion. Road-region background correspondences provide a geometric reference for estimating camera motion in driving videos. The residual motion features are then matched against language expressions to yield a matching score, which is integrated with the host model's prediction score via late fusion. On the Refer-KITTI datasets, the pooled HOTA improves for different host models. On the moving-class expressions, the largest gain is observed when iKUN is used as the host model ($+9.44\pm0.92$).

Despite these promising results, several limitations remain. First, the road-plane homography assumes a flat ground plane. Scenes with steep slopes or uneven terrain fall outside the model's scope. Second, the keyword classifier routes only moving-class expressions to $\alpha_{\mathrm{mot}}$; direction and turning expressions ($214$ of the $818$ in Refer-KITTI V1) receive the other-class weight $\alpha_{\mathrm{oth}}$, so the module does not help them even though they describe motion. Finally, our experiments are conducted on three RMOT frameworks but only on the Refer-KITTI benchmarks. Additional evaluations on other datasets are needed to further validate the generality of the proposed approach.

\bibliographystyle{IEEEbib}
\bibliography{refs}

@inproceedings{transrmot,
  title     = {Referring Multi-Object Tracking},
  author    = {Wu, Dongming and Han, Wencheng and Wang, Tiancai and Dong, Xingping and Zhang, Xiangyu and Shen, Jianbing},
  booktitle = {Proceedings of the IEEE/CVF Conference on Computer Vision and Pattern Recognition (CVPR)},
  publisher = {IEEE},
  pages     = {14633--14642},
  year      = {2023},
  doi       = {10.1109/CVPR52729.2023.01406}
}

@inproceedings{ikun,
  title     = {{iKUN}: Speak to Trackers without Retraining},
  author    = {Du, Yunhao and Lei, Cheng and Zhao, Zhicheng and Su, Fei},
  booktitle = {Proceedings of the IEEE/CVF Conference on Computer Vision and Pattern Recognition (CVPR)},
  publisher = {IEEE},
  pages     = {19135--19144},
  year      = {2024},
  doi       = {10.1109/CVPR52733.2024.01810}
}

@misc{flexhook,
  title         = {Rethinking Two-Stage Referring-by-Tracking in Referring Multi-Object Tracking: Make it Strong Again},
  author        = {Li, Weize and Du, Yunhao and Yin, Qixiang and Zhao, Zhicheng and Su, Fei},
  year          = {2025},
  eprint        = {2503.07516},
  archivePrefix = {arXiv},
  primaryClass  = {cs.CV},
  doi           = {10.48550/arXiv.2503.07516},
  note          = {Accepted to CVPR 2026}
}

@misc{temprmot,
  title         = {Bootstrapping Referring Multi-Object Tracking},
  author        = {Zhang, Yani and Wu, Dongming and Han, Wencheng and Dong, Xingping},
  year          = {2024},
  eprint        = {2406.05039},
  archivePrefix = {arXiv},
  primaryClass  = {cs.CV},
  doi           = {10.48550/arXiv.2406.05039},
  note          = {Introduces TempRMOT and Refer-KITTI-V2}
}

@misc{vmrmot,
  title         = {Vision--Motion--Reference Alignment for Referring Multi-Object Tracking via Multi-Modal Large Language Models},
  author        = {Lv, Weiyi and Zhang, Ning and Sun, Hanyang and Jiang, Haoran and Zhao, Kai and Xiao, Jing and Zeng, Dan},
  year          = {2025},
  eprint        = {2511.17681},
  archivePrefix = {arXiv},
  primaryClass  = {cs.CV},
  doi           = {10.48550/arXiv.2511.17681}
}

@misc{botsort,
  title         = {{BoT-SORT}: Robust Associations Multi-Pedestrian Tracking},
  author        = {Aharon, Nir and Orfaig, Roy and Bobrovsky, Ben-Zion},
  year          = {2022},
  eprint        = {2206.14651},
  archivePrefix = {arXiv},
  primaryClass  = {cs.CV},
  doi           = {10.48550/arXiv.2206.14651}
}

@inproceedings{ucmctrack,
  title     = {{UCMCTrack}: Multi-Object Tracking with Uniform Camera Motion Compensation},
  author    = {Yi, Kefu and Luo, Kai and Luo, Xiaolei and Huang, Jiangui and Wu, Hao and Hu, Rongdong and Hao, Wei},
  booktitle = {Proceedings of the AAAI Conference on Artificial Intelligence},
  publisher = {AAAI Press},
  volume    = {38},
  pages     = {6702--6710},
  year      = {2024},
  doi       = {10.1609/aaai.v38i7.28493}
}

@article{hota,
  title   = {{HOTA}: A Higher Order Metric for Evaluating Multi-Object Tracking},
  author  = {Luiten, Jonathon and Os{\v{e}}p, Aljo{\v{s}}a and Dendorfer, Patrick and Torr, Philip and Geiger, Andreas and Leal-Taix{\'e}, Laura and Leibe, Bastian},
  journal = {International Journal of Computer Vision},
  volume  = {129},
  number  = {2},
  pages   = {548--578},
  year    = {2021},
  doi     = {10.1007/s11263-020-01375-2}
}

@article{ransac,
  title   = {Random Sample Consensus: A Paradigm for Model Fitting with Applications to Image Analysis and Automated Cartography},
  author  = {Fischler, Martin A. and Bolles, Robert C.},
  journal = {Communications of the ACM},
  volume  = {24},
  number  = {6},
  pages   = {381--395},
  year    = {1981},
  doi     = {10.1145/358669.358692}
}

@inproceedings{sbert,
  title     = {{Sentence-BERT}: Sentence Embeddings using Siamese {BERT}-Networks},
  author    = {Reimers, Nils and Gurevych, Iryna},
  booktitle = {Proceedings of the 2019 Conference on Empirical Methods in Natural Language Processing and the 9th International Joint Conference on Natural Language Processing (EMNLP-IJCNLP)},
  pages     = {3982--3992},
  year      = {2019},
  address   = {Hong Kong, China},
  publisher = {Association for Computational Linguistics},
  doi       = {10.18653/v1/D19-1410}
}

@misc{infonce,
  title         = {Representation Learning with Contrastive Predictive Coding},
  author        = {van den Oord, Aaron and Li, Yazhe and Vinyals, Oriol},
  year          = {2018},
  eprint        = {1807.03748},
  archivePrefix = {arXiv},
  primaryClass  = {cs.LG}
}

@inproceedings{deeprmot,
  title        = {Visual-Linguistic Representation Learning with Deep Cross-Modality Fusion for Referring Multi-Object Tracking},
  author       = {He, Wenyan and Jian, Yajun and Lu, Yang and Wang, Hanzi},
  booktitle    = {ICASSP 2024 -- 2024 IEEE International Conference on Acoustics, Speech and Signal Processing (ICASSP)},
  pages        = {6310--6314},
  year         = {2024},
  organization = {IEEE},
  doi          = {10.1109/ICASSP48485.2024.10447535}
}

@misc{emap,
  title         = {Ego-Motion Aware Target Prediction Module for Robust Multi-Object Tracking},
  author        = {Mahdian, Navid and Jani, Mohammad and Soufi Enayati, Amir M. and Najjaran, Homayoun},
  year          = {2024},
  eprint        = {2404.03110},
  archivePrefix = {arXiv},
  primaryClass  = {cs.CV}
}

@misc{refergpt,
  title         = {ReferGPT: Towards Zero-Shot Referring Multi-Object Tracking},
  author        = {Chamiti, Tzoulio and Di Bella, Leandro and Munteanu, Adrian and Deligiannis, Nikos},
  year          = {2025},
  eprint        = {2504.09195},
  archivePrefix = {arXiv},
  primaryClass  = {cs.CV},
  note          = {CVPR Workshops}
}

@misc{mex,
  title         = {MEX: Memory-efficient Approach to Referring Multi-Object Tracking},
  author        = {Tran, Huu-Thien and Pham, Phuoc-Sang and Tran, Thai-Son and Luu, Khoa},
  year          = {2025},
  eprint        = {2502.13875},
  archivePrefix = {arXiv},
  primaryClass  = {cs.CV}
}

@inproceedings{shitomasi,
  title     = {Good Features to Track},
  author    = {Shi, Jianbo and Tomasi, Carlo},
  booktitle = {IEEE Conference on Computer Vision and Pattern Recognition (CVPR)},
  pages     = {593--600},
  year      = {1994}
}

@inproceedings{lucaskanade,
  title     = {An Iterative Image Registration Technique with an Application to Stereo Vision},
  author    = {Lucas, Bruce D. and Kanade, Takeo},
  booktitle = {International Joint Conference on Artificial Intelligence (IJCAI)},
  pages     = {674--679},
  year      = {1981}
}

@inproceedings{adamw,
  title     = {Decoupled Weight Decay Regularization},
  author    = {Loshchilov, Ilya and Hutter, Frank},
  booktitle = {International Conference on Learning Representations (ICLR)},
  year      = {2019}
}

@inproceedings{storm,
  title     = {{STORM}: End-to-End Referring Multi-Object Tracking in Videos},
  author    = {Lu, Zijia and Yi, Jingru and Wang, Jue and Chen, Yuxiao and Chen, Junwen and Li, Xinyu and Modolo, Davide},
  booktitle = {Proceedings of the IEEE/CVF Conference on Computer Vision and Pattern Recognition (CVPR), Findings},
  year      = {2026}
}

\end{document}